\relax
\documentclass[letterpaper]{article} 
\usepackage{aaai20}  
\usepackage{times}  
\usepackage{helvet} 
\usepackage{courier}  
\usepackage[hyphens]{url}  
\usepackage{graphicx} 
\usepackage{graphicx}  
\usepackage{amsfonts,amssymb}
\usepackage {mathtools}
\usepackage{multirow}
\usepackage{floatrow}
\usepackage{booktabs,subcaption,dcolumn}
\usepackage{xcolor}
\definecolor{mygray}{gray}{0.6}

\title{TEINet: Towards an Efficient Architecture for Video Recognition}

\author{
\begin{tabular}{cccccc}
\multicolumn{1}{c}{Zhaoyang Liu\textsuperscript{\rm 1}\thanks{ indicates equal contribution.} \quad Donghao Luo\textsuperscript{\rm 2}\footnotemark[1] \quad Yabiao Wang\textsuperscript{\rm 2} \quad Limin Wang\textsuperscript{\rm 1}\thanks{Corresponding author (lmwang@nju.edu.cn).}} \\
\multicolumn{1}{c}{Ying Tai\textsuperscript{\rm 2} \quad Chengjie Wang\textsuperscript{\rm 2} \quad Jilin Li\textsuperscript{\rm 2} \quad Feiyue Huang\textsuperscript{\rm 2} \quad Tong Lu\textsuperscript{\rm 1}}\\
\end{tabular} \\ \\
\textsuperscript{\rm 1} State Key Lab for Novel Software Technology, Nanjing University, China \\
\textsuperscript{\rm 2} Youtu Lab, Tencent \\
}

\begin{document}

\maketitle

\begin{abstract}
Efficiency is an important issue in designing video architectures for action recognition.
3D CNNs have witnessed remarkable progress in action recognition from videos. However, compared with their 2D counterparts, 3D convolutions often introduce a large amount of parameters and cause high computational cost.
To relieve this problem, we propose an efficient temporal module, termed as {\em Temporal Enhancement-and-Interaction} (\textbf{TEI Module}), which could be plugged into the existing 2D CNNs (denoted by TEINet).
The TEI module presents a different paradigm to learn temporal features by decoupling the modeling of channel correlation and temporal interaction. 
First, it contains a {\em Motion Enhanced Module} (\textbf{MEM}) which is to enhance the motion-related features while suppress irrelevant information (e.g., background).
Then, it introduces a {\em Temporal Interaction Module} (\textbf{TIM}) which supplements the temporal contextual information in a channel-wise manner.
This two-stage modeling scheme is not only able to capture temporal structure flexibly and effectively, but also efficient for model inference.
We conduct extensive experiments to verify the effectiveness of TEINet on several benchmarks (e.g., Something-Something V1\&V2, Kinetics, UCF101 and HMDB51). Our proposed TEINet can achieve a good recognition accuracy on these datasets but still preserve a high efficiency.
\end{abstract}

\begin{figure}
  \centering
  \includegraphics[width=8cm]{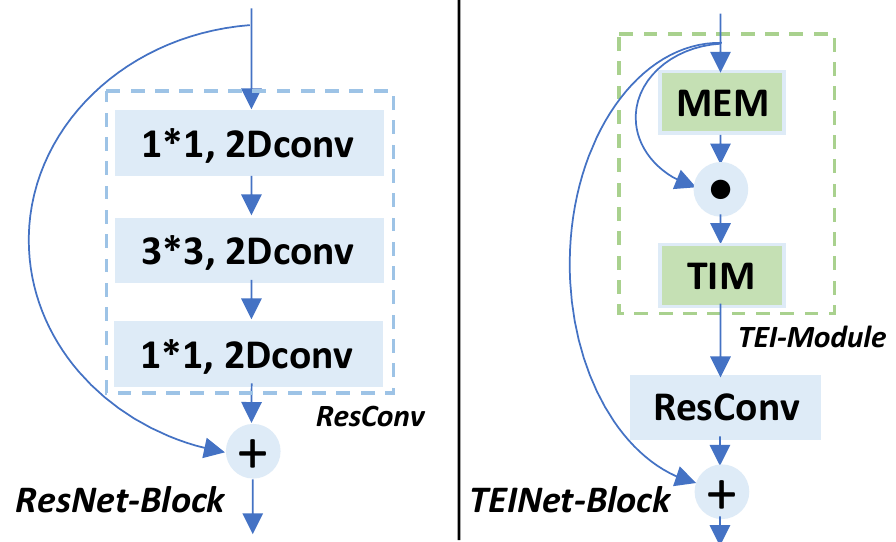}
  \caption{{\bf TEINet building block}. We present an effective TEI module to decouple temporal modeling into a MEM to enhance motion-related features and a TIM capture 
  temporal contextual information. This TEI module could be inserted into the 2D ResNet block to construct an efficient video architecture of TEINet.}
  \label{fig:overall_arch}
\end{figure}

\section{Introduction}
Video understanding is one of the most important problems in computer vision~\cite{two_stream,3d_conv,tsn}. Action recognition is a fundamental task in video understanding, as it is able to not only extract semantic information from videos, but also yield general video representations for other tasks such as action detection and localization~\cite{slowfast,ssn}. Unlike static images, the core problem of action recognition is how to model temporal information effectively. Temporal dimension typically exhibits a different property with respect to spatial domain. Modeling temporal information in a proper way is crucial for action recognition, which has aroused great interest of research.

Recently the convolutional networks~\cite{lecun-98} have become the mainstream method in action recognition~\cite{two_stream,I3D,r(2+1)d}. TSN~\cite{tsn} is an efficient method which ignores the constraint on temporal order information and only aggregates temporal information at the final classifier layer. To capture temporal information slowly and earlier, some new efficient 2D CNN based architectures are developed such as StNet~\cite{stnet} and TSM~\cite{tsm}. However, they involve some hand-crafted designs, which lacks a clear explanation and may be suboptimal for temporal modeling. 3D CNNs~\cite{3d_conv,I3D} are more principled architectures for directly learning spatiotemporal features from RGB frames. Unfortunately, this simple extension from a 2D convolution to its 3D version leads to a critical issue: it causes high computational cost when densely replacing 2D convolutions with 3D counterparts. Therefore we hope to devise a flexible temporal modeling module which shares the capacity of learning spatiotemporal representations yet still keeps the efficiency of 2D CNNs.

Intuitively, temporal structure in video can benefit action recognition from multiple aspects. Firstly, motion information is able to help us focus on moving objects or people that are discriminative for action recognition. These discriminative features could be automatically determined for each input video. Secondly, the temporal evolution of visual features enables us to capture dynamic variation in videos and relate adjacent frame-level features for action recognition. Based on these analyses, we propose a new temporal modeling paradigm, termed as Enhance-and-Interact. This new design decouples the temporal module into two stages: first enhance discriminative features and then capture their temporal interaction. This unique design enables us to separately capture the channel-level correlation and temporal relation in a more principled and efficient way. It turns out that this separate modeling scheme is able to not only capture temporal structure flexibly and effectively, but also keeps a high inference efficiency in practice.

Specifically, we first present the {\em Motion Enhanced Module} (MEM), which utilizes motion information as a guide to focus on important features. To make this enhancement more efficient and effective, we squeeze the feature maps to only focus on channel-level importance and exploit temporal difference as an approximate motion map. Then, to capture the temporal interaction of adjacent frames, we present the {\em Temporal Interaction Module} (TIM), which model the local temporal variations of visual features. To control the model complexity and ensure the inference efficiency, we employ a temporal channel-wise convolution in a local time window. These two modules are plugged sequentially to yield a novel temporal module, namely {\em Temporal Enhancement-and-Interaction} (TEI module), which is {\bf a generic building block and could be plugged into the existing 2D CNNs such as ResNets}, as illustrated in Figure~\ref{fig:overall_arch}.

In experiments, we verify the effectiveness of TEI block with the 2D ResNet on the large-scale datasets such as Kinetics~\cite{kinetics} and Something-Something~\cite{sth}. The final video architecture, coined as TEINet, obtains an evident performance improvement over previous approaches while is still able to keep fast inference speed. In particular, our TEINet achieves the state-of-the-art performance on the dataset of Something-Something, and comparable performance to previous 3D CNN based methods at a lower computational cost on the dataset of Kinetics. We also demonstrate the generalization ability of TEINet by fine-tuning on the datasets of UCF101 and HMDB51, where competitive recognition accuracy is also obtained. The main contribution in this work is summarized as follows:
\begin{itemize}
    \item We present a new temporal modeling module, termed as TEI module, by decoupling the task of temporal feature learning into channel-level enhancement and local temporal interaction.
    \item The proposed TEINet is verified on various large-scale datasets, demonstrating that it is able to obtain an evident improvement over previous temporal modeling methods with a lower computational cost.
    
\end{itemize}

\begin{figure*}
  \centering
  \includegraphics[width=18cm]{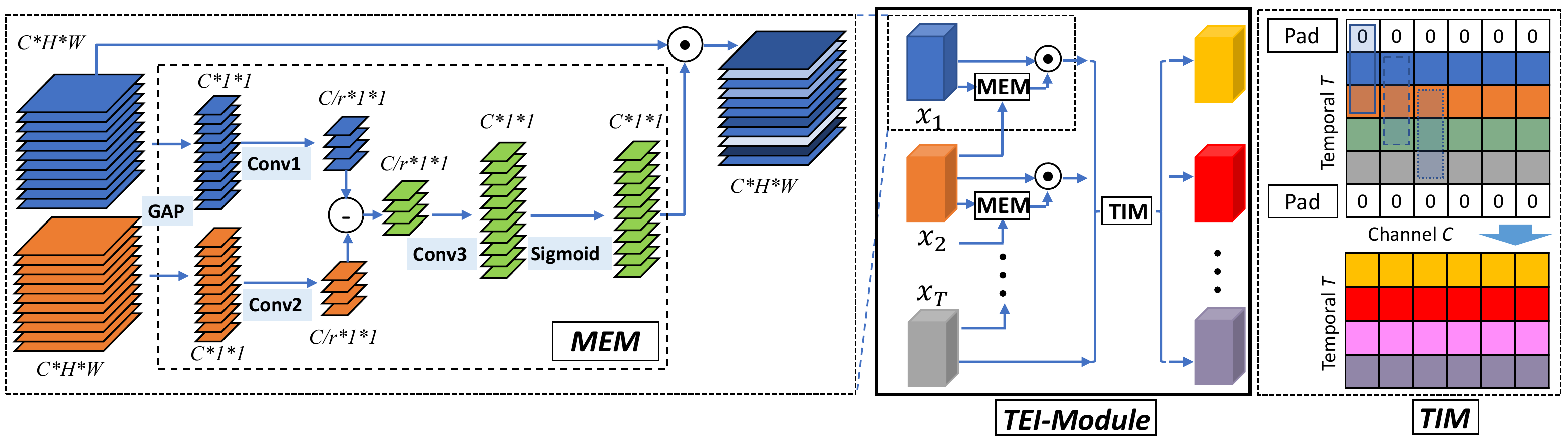}
  \caption{The pipeline of TEI module. We show motion enhanced module (\textbf{MEM}) in the left and temporal interaction module (\textbf{TIM}) in the right. The $\odot$ denotes element-wise multiplication, and $\ominus$ denotes element-wise subtraction. Notably, in TIM, we use different box to represent kernel weights, which means each channel do not share kernel weights.}
  \label{fig:mem_tim}
\end{figure*}

\section{Related Work}

\noindent\textbf{2D CNNs in Action Recognition.}
Conventional 2D CNNs were extensively applied on action recognition in videos~\cite{two_stream,st_resnet,tsn,tsm,gan2015devnet}. Two stream methods~\cite{two_stream,two_stream_fusion,ZhangWW0W16} regarded optical flow or motion vector as motion information to make up a temporal stream CNN.
TSN~\cite{tsn} utilized average pooling to aggregates temporal information from set of sparsely-sampled frames.
To improve the temporal reasoning ability of TSN, the TRN~\cite{trn} was proposed by focusing on the multi-scale temporal relations among sampled frames.
To model temporal structure efficiently, TSM~\cite{tsm} proposed a temporal shift module on the original feature map. 
Sharing the same motivation with TSM, our TEINet is also based on 2D backbones with high efficiency, but better at capturing temporal clues for video recognition.

\noindent\textbf{3D CNNs in Action Recognition.}
3D convolution~\cite{3d_conv,I3D} was a straightforward extension over 2D versions to learn the spatiotemporal representation directly from RGB. I3D~\cite{I3D} inflated all 2D convolution kernels into 3D convolution kernels and directly utilized the pre-trained weights on ImageNet~\cite{imagenet}. ARTNet~\cite{artnet} improved the original 3D convolutions with higher-order relation modeling to explicitly capture motion.
3D convolution is natural and simple way for modeling temporal features, yet in practice with heavy computation. Unlike 3D CNNs, our TEINet resorts to a new temporal module purely based on 2D CNNs for video recognition.

\noindent\textbf{Efficient Temporal Modules.}
Some efficient temporal models were proposed by using combination of 2D and 3D convolutions. 
ECO~\cite{eco} combined the 2D convolution and 3D convolution into one network to achieve a balance between 2D CNNs and 3D CNNs.
To decompose the optimization of spatial and temporal features, pseudo-3D convolution, e.g., P3D~\cite{p3d}, S3D~\cite{s3d} and R(2+1)D~\cite{r(2+1)d}, decomposed the spatio-temporal 3D convolution into a spatial 2D convolution and a temporal 1D convolution.
Our TEINet integrates a new temporal block into a purely 2D backbone to endow network with the ability to model temporal structure in videos.

\noindent\textbf{Attention in Action Recognition.}
The attention mechanism~\cite{senet,sge} were widely used in image classification, which can boost performance using a small portion of extra parameters. Similarly, there are some works~\cite{nonlocal,att_pooling}
related to attention in action recognition. Non-local network formulates the non-local mean operation as Non-local block to capture the long-range dependencies in video.
Motion enhanced module (MEM) in our method differs from these attention methods. MEM constructs the temporal attention weights by local motion tendency which can be trained by end-to-end without using extra supervision and give a sizable boost in accuracy.

\section{Method}
\label{method}
We will introduce our proposed TEI module in this section. First, we describe motion enhanced module and explain how to learn a channel level attention weights. Then we present the technical details of temporal interaction module. Finally we combine these two modules as a TEINet building block and integrate this block into the off-the-shelf architecture of 2D CNN.

\subsection{Motion Enhanced Module}
\label{method:MEM}

In action recognition, spatial features can only provide partial information for action recognition. It is well established that motion information is an crucial cue for understanding human behavior in videos. Consequently, we first design a {\em Motion Enhanced Module} (\textbf{MEM}) to focus on motion-salient features while suppress the irrelevant information at background.

Our method is to enhance the motion-related features in a channel-wise way by using the temporal difference of adjacent frame level features. To decrease the computational cost, we first construct a global representation for each channel and then perform feature enhancement in a channel level. As depicted in Figure~\ref{fig:mem_tim}, given an input sequence $X=\{x_1,x_2,...,x_T\}$ in which $x_t$ $\in \mathbb{R}^{C \times H \times W}$, We first aggregate feature map $x_t$ across their spatial dimensions ($H\times W$) by using global average pooling, which produces $\hat{x_t} \in \mathbb{R}^{C \times 1 \times 1}$. Then, these pooled features go through subsequent processing operations to generate the channel importance weights.

Basically, we observe that the overall appearance information varies gradually and slowly over time. The pixel values in motion salient regions would change more quickly than those in static regions. In practice, we exploit the feature difference between adjacent frames to approximately represent the motion saliency.
To reduce the model complexity, the $\hat{x_t}$ and $\hat{x}_{t+1}$ are fed into two different 2D convolutions whose kernel size of $1\times 1$ in which the channels of $\hat{x_t}$ will be compressed. This dimension reduction and difference calculation can be formulated as:
\begin{equation}
\begin{aligned}
s_t&=\mathrm{Conv1}(\hat{x_t}, W_\theta) - \mathrm{Conv2}(\hat{x}_{t+1}, W_\phi) .
\end{aligned}
\label{methods:reduce_conv1}
\end{equation}
Here $W_\theta$ and $W_\phi$ are learnable parameters of the convolutions that reduce the number of channels in  $\hat{X}$ from $C$ to $\frac{C}{r}$. In our experiments, the reduction ratio $r$ is set to 8.

Then another 2D convolution is applied on $s_t$, which aims to recover the channel dimension of $s_t$ as same as input $s_t$. The attention weights are obtained by:
\begin{equation}
\begin{aligned}
\hat{s_t}=\sigma(\mathrm{Conv3}(s_t, W_\varphi)),
\end{aligned}
\label{methods:reduce_conv2}
\end{equation}
where $\sigma(*)$ denotes a sigmoid function and $W_\varphi$ are learnable parameters of $\mathrm{Conv3}$.
Finally we obtain attention weights $\hat{s}\in \mathbb{R}^{ C \times 1 \times 1} $ for different channels. We utilize channel-wise multiplication to enhance motion-salient features:
\begin{equation}
\begin{aligned}
u_t = \hat{s_t} \cdot x_t ,
\end{aligned}
\label{methods:reduce_conv2}
\end{equation}
where $t \in [1, T-1]$ and $u_t$ is our final enhanced feature map. To keep the temporal scale consistent with input $X$, we simply copy the $x_T$ as $u_T$, namely, $u_T=x_T$.

{\bf Discussion.} We have noticed that our MEM is similar to SE module in ~\cite{senet}. However, the essential difference between SE module and MEM is that SE module is a kind of self attention mechanism by using its own global feature to calibrate the different channels, while our MEM is a motion-aware attention module by enhancing the motion-related features. To prove the effectiveness of MEM, we conduct the comparative experiments in section~\ref{ablation_study}. Under the same setting, our MEM is better at enhancing temporal features for action recognition than SE module in video dataset.

\subsection{Temporal Interaction Module}
\label{method:tim}

In MEM, we enhance the motion-related features, but our model is still incapable of capturing temporal information in a local time window, namely the temporal evolution of visual pattern over time. Consequently, we propose the {\em Temporal Interaction Module} (TIM) which aims to capture temporal contextual information at a low computational cost.
More specifically, we here use a channel-wise convolution to learn the temporal evolution for each channel independently, which preserves low computational complexity for model design.

As illustrated in Figure\ref{fig:mem_tim}, given a input $U=\{u_1, u_2,...,u_T\}$, we first transform its shape from $U^{T \times C \times H \times W}$ to $\hat{U}^{C \times T \times H\times W}$ (denoted by $\hat{U}$ to avoid ambiguity).
Then we apply the channel-wise convolution to operate on $\hat{U}$ as follows:
\begin{equation}
\begin{aligned}
Y_{c,t,x,y} = \sum_{i} V_{c,i} \cdot \hat{U}_{c,t+i,x,y} ,
\end{aligned}
\label{methods:tim_conv}
\end{equation}
where $V$ is the channel-wise convolutional kernel and $Y_{c,t,x,y}$ is the output after temporal convolution.
The channel-wise convolution tremendously decreases the computation costs comparing with 3D convolution.
In our setting, the kernel size of the channel-wise convolution is $3\times 1 \times 1$, which implies the features are only interacting with features in adjacent time, but the temporal receptive fields will gradually grow when feature maps pass through deeper layers of network. After convolution, we will transform the shape of output $Y$ back to $T \times C \times H \times W$. The parameters of  vanilla 3D convolution is $C_{out}\times C_{in}\times t \times d \times d$, and the temporal 1D convolution in \cite{r(2+1)d} is $C_{out}\times C_{in}\times t$, but the parameters of TIM is $C_{out}\times 1 \times t$. The number of parameters in TIM is greatly reduced when compared with other temporal convolutional operators. 

{\bf Discussion.} We figure out our TIM is related to the recent proposed TSM~\cite{tsm}. In fact, TSM could be viewed as a channel-wise temporal convolution, where temporal kernel is fixed as $[0,1,0]$ for non shift, $[1,0,0]$ for backward shift, and $[0,0,1]$ for forward shift. Our TIM generalizes the TSM operation into a flexible module with a learnable convolutional kernel. In experiment, we find that this learnable scheme is more effective than random shift to capture temporal contextual information for action recognition.

\subsection{TEINet}
After introducing the MEM and TIM, we are ready to describe how to build the temporal enhancement-and-interaction block (TEI) and integrate it into the existing network architecture. As shown in Figure~\ref{fig:overall_arch}, the TEI module is composed of MEM and TIM introduced above, which could be implemented efficiently.
First the input feature maps will be fed into MEM to learn attention weights for different channels, which aims to enhance the motion-related features.
Then the enhanced features will be fed into the TIM to capture temporal contextual information.
Our TEI module is a generic and efficient temporal modeling module, which could be plugged into any existing 2D CNN to capture temporal information, and the resulted network is called {\em Temporal Enhancement-and-Interaction Network} (TEINet). 

Our TEI module is directly inserted into the 2D CNN backbone, while other methods~\cite{r(2+1)d,p3d,s3d} replace the 2D convolutions with more expensive 3D convolutions or (2+1)D convolutions. This new integration method is able to not only use the pre-trained ImageNet model for initialization but also bring a smaller number of extra computational FLOPs compared with 3D CNNs.
In our experiments, to trade off between performance and computational cost, we instantiate the temporal enhancement-and-interaction network (TEINet) using ResNet-50~\cite{resnet} as backbone. We conduct extensive experiments to figure out the optimal setting of TEINet for action recognition in Section.~\ref{experiments}. 

{\bf Discussion.} Our paper proposed enhancement-and-interaction is a factorized modeling method to endow network with a strong ability to learn the temporal features in videos. We find that our module is effective for both types of video datasets: motion dominated one such as Something-Something V1\&V2 and appearance dominated one such as Kinetics-400. MEM and TIM focus on different aspects when capturing temporal information, where MEM aims to learn channel level importance weights and TIM tries to learn temporal variation pattern of adjacent features. These two modules are cooperative and complementary to each other as demonstrated in Table~\ref{tab:impact_of_two_modules}. 

\begin{table*}
\centering
	\begin{subtable}[t]{0.31\textwidth}
		\centering
		\begin{tabular}{ccc}
         \toprule[1pt]
         model & Top-1 & Top-5\\
         \hline
         Res50+TSN & 19.7\% & 46.6\% \\
         Res50+TSM & 43.4\% & 73.2\% \\
         \hline
         Res50+MEM & 33.5\% & 61.5\%\\
         Res50+TIM & 46.1\% & 74.7\%\\
         Res50+SE+TIM &46.1\% & 75.2\%\\
         Res50+MEM+TIM & \textbf{47.4\%} & \textbf{76.6\%}\\
         \hline
         \end{tabular}
		\caption{Exploration on MEM and TIM, and comparison with other baseline methods.}
		\label{tab:impact_of_two_modules}
	\end{subtable}
	\quad \quad \quad
	\begin{subtable}[t]{0.22\textwidth}
		\centering
		\begin{tabular}{ccc}
         \toprule[1pt]
         stage  &Top-1 & Top-5\\
         \hline
         res$_2$  & 41.6\% & 70.1\% \\
         res$_3$ & 43.1\% & 72.1\% \\
         res$_4$ & \textbf{45.4\%} & \textbf{74.6\%}\\
         res$_5$ & 45.3\% & 74.3\%\\
         \hline
         \end{tabular}
         \caption{The TEI blocks in different stage of ResNet-50}
		\label{tab:insert_block}
	\end{subtable}
	\quad \quad
	\begin{subtable}[t]{0.3\textwidth}
		\centering
		\begin{tabular}{cccc}
         \toprule[1pt]
         stages & Blocks & Top-1 & Top-5\\
         \hline
         res$_5$ & 3 & 45.3\% & 74.3\%\\
         res$_{4-5}$ & 9 & 46.7\%& 76.3\% \\
         res$_{3-5}$ & 13 & 47.3\% & 75.2\%\\
         res$_{2-5}$ & 16 & \textbf{47.4\%} & \textbf{75.8\%}\\
         \hline
         \end{tabular}
		\caption{The number of TEI block inserted into ResNet-50.}
		\label{tab:number_of_block}
	\end{subtable}
	\caption{Ablation studies on Something-Something V1.}
	\label{tab:studies_on_TEINet}
\end{table*}

\begin{table*}[th]
\centering
\scalebox{0.9}{
\begin{tabular}{ccccccc}
\toprule[2pt]
Method  & Frame & Params & FLOPs & Latency & Throughput & Sthv1 \\
\hline
I3D~(Carreira et al. 2017)  & 64 & 35.3M & 360G & 165.3ms & 6.1 vid/s & 41.6\% \\

ECO$_{16f}$~(Zolfaghari et al. 2018)  &16 & 47.5M  & 64G & 30.6ms & 45.6 vid/s &41.4\% \\
TSN~\cite{tsn} & 8 & 24.3M & 33G & 15.5ms & 81.5 vid/s & 19.7\% \\
TSM~(Lin et al. 2018)  & 8 & 24.3M & 33G & 17.4ms & 77.4 vid/s & 43.4\% \\
TSM~(Lin et al. 2018)  & 16 & 24.3M & 65G & 29.0ms & 39.5 vid/s & 44.8\% \\

\hline
Res50+TIM & 8 & 24.3M & 33G & 20.1ms & 61.6 vid/s& 46.1\% \\
TEINet & 8 & 30.4M & 33G & 36.5ms & 46.9 vid/s & 47.4\%\\
Res50+TIM & 16 & 24.3M & 66G &  34.9ms & 31.4 vid/s & 48.5\% \\
TEINet & 16 & 30.4M & 66G & 49.5ms & 24.2 vis/s & 49.9\%\\
\bottomrule[2pt]

\end{tabular}
}
\caption{Quantitatively analysis on latency and throughput Something-Something V1.  "vid/s" represents videos per second. The larger latency and the smaller throughput represent higher efficiency.}
\label{tab:runtime}
\end{table*}

\section{Experiments}
\label{experiments}
\subsection{Datasets}

\textbf{Something-Something V1\&V2.} ~\cite{sth} is a large collection of video clips containing daily actions interacting with common objects. It tries to focuses on motion itself without differentiating manipulated objects. V1 includes 108499 video clips, and V2 includes 220847 video clips.
They both have 174 classes.
 
\noindent \textbf{Kinetics-400.} ~\cite{kinetics} is a large-scale dataset in action recognition, which contains 400 human action classes, with at least 400 video clips for each class. Each clip is collected from YouTube videos and then trimmed to around 10s. The newest version of Kinetics has updated to Kinetics-700 which approximately includes 650k video clips that covers 700 human action classes. For fair comparison with previous methods, we conduct experiments on Kinetics-400.
 
\noindent {\bf UCF101 and HMDB51.} Finally, to verify the generalization ability to transfer to smaller scale datasets, we report the results on the datasets of UCF101 ~\cite{ucf101} and HMDB51~\cite{hmdb51}. The UCF101 contains 101 categories with around 13k videos, while HMDB51 has about 7k videos spanning over 51 categories. On UCF101 and HMDB51, we follow the common practice that reports the accuracy by averaging over three splits.
Different from Something-Something, the datasets of Kinetics-400, UCF101 and HMDB51 are less sensitive to temporal relationship.

\subsection{Experimental Setup}
We here choose the ResNet-50 as our backbone for the trade off between performance and efficiency. Unless specified, our model is pre-trained on ImageNet~\cite{imagenet}.

\noindent \textbf{Training.}
We applied a similar pre-processing method to~\cite{nonlocal}: first resizing the shorter side of raw images to 256 and then employing a center cropping and scale-jittering. Before being fed into the network, the images will be resized to $224\times224$. In our model, We attempt to stack 8 frames or 16 frames as a clip.
On the Kinetics dataset, we train our models for 100 epochs in total, starting with a learning rate of 0.01 and reducing to its $\frac{1}{10}$ at 50, 75, 90 epochs.
For fair comparisons with the state-of-the-art models, we follow the testing strategy in \cite{tsm}, which uniformly samples 8 or 16 frames from the consecutive 64 frames randomly sampled in each video.
We observe that the duration of most videos in Something-Something V1\&V2 normally has less than 64 frames. Thus we use the similar strategy to TSN~\cite{tsn} to train our model. Specifically, we uniformly sample the 8 or 16 frames from each video. On Something-Something V1\&V2, We train the TEINet for 50 epochs starting with a learning rate 0.01 and reducing it by a factor of 10 at 30, 40, 45 epochs.
For all of our experiments, we utilize SGD with momentum 0.9 and weight decay of 1e-4 to train our TEINet on Tesla M40 GPUs using a mini batch size of 64.

\noindent \textbf{Inference.}
We follow the widely used settings in \cite{nonlocal,tsm}: resizing shorter side to 256 and taking 3 crops (left, middle, right) in each frame.
Then we uniformly sample 10 clips in each video and compute the classification scores for all clips individually. The final prediction will be obtained by utilizing the average pooling to aggregate the scores of 10 clips.

\begin{table*}[th]\small
\centering
\scalebox{0.95}{
\begin{tabular}{c|c|c|c|c|c|c}
\toprule[2pt]
\multirow{2}*{Method} & \multirow{2}*{Backbone} & \multirow{2}*{Pre-train} &\multirow{2}*{Frames} & \multirow{2}*{FLOPs} &Val& Test \\
&  &  &  & &Top-1 &Top-1\\
\hline
TSN-RGB~\cite{tsn}& ResNet2D-50 &ImgNet & $8f$ &33G & 19.7\% & -\\
\hline \hline
TRN-Multiscale-RGB~\cite{trn}& BNInception &\multirow{3}*{ImgNet} & $8f$ & 33G & 34.4\% & 33.6\%\\
TRN-Multiscale-RGB~\cite{trn}& ResNet2D-50 & & $8f$ & 33G & 38.9\% & -\\
\color{mygray}TRN-Multiscale-2Stream~\cite{trn}& \color{mygray}BNInception & & \color{mygray}$8f+8f$ & \color{mygray}- & \color{mygray}42.0\% & \color{mygray}40.7\%\\
\hline\hline
S3D-G-RGB~\cite{s3d} & Inception &ImgNet & $64f$ & 71G & 48.2\% & - \\
\hline\hline
I3D-RGB~\cite{i3d_gcn}& ResNet3D-50 & \multirow{3}*{ImgNet+K400}& \multirow{3}*{$32f\times2$} & 306G & 41.6\% & -\\
NL I3D-RGB~\cite{i3d_gcn}& ResNet3D-50 & & & 334G & 44.4\% & -\\
NL I3D+GCN-RGB~\cite{i3d_gcn}& ResNet3D-50+GCN & & & 606G & 46.1\% & 45.0\%\\
\hline\hline
ECO-RGB~(Zolfaghari et al. 2018) & \multirow{3}*{BNIncep+Res3D-18} & \multirow{3}*{K400} & $16f$ & 64G & 41.6\% & -\\
ECO-RGB~(Zolfaghari et al. 2018) & & & $92f$ & 267G & 46.4\% & -\\
\color{mygray}ECO$_{En}$Lite-2Stream~(Zolfaghari et al. 2018) &  & & \color{mygray}$92f+92f$ & - & \color{mygray}49.5\% & \color{mygray}43.9\\
\hline\hline
TSM-RGB~\cite{tsm} & \multirow{4}*{ResNet2D-50} & \multirow{4}*{ImgNet+K400}& $8f$ & 33G & 43.4\% & -\\
TSM-RGB~\cite{tsm} &  & & $16f$& 65G & 44.8\% & -\\
TSM$_{En}$-RGB~\cite{tsm} &  & & $16f+8f$& 98G & 46.8\% & -\\
\color{mygray}TSM-2Stream~\cite{tsm} &  & & \color{mygray}$16f+16f$ & - & \color{mygray}50.2\% & \color{mygray}47.0\\
\hline
\hline
\multirow{4}*{TEINet-RGB} & \multirow{4}*{ResNet2D-50} & \multirow{4}*{ImgNet} & $8f$ & 33G & \textbf{47.4\%} & -\\
 &  & & $8f\times 10$ & 990G & 48.8\% & -\\
 &  & & $16f$ & 66G & \textbf{49.9\%} & -\\
 &  & &  $16f\times 10$ & 1980G & 51.0\% & 44.7\%\\
\hline
TEINet$_{En}$-RGB & ResNet2D-50 & ImgNet &  $16f+8f$ & 99G & \textbf{52.5}\% & \textbf{46.1}\%\\
\bottomrule[2pt]
\end{tabular}
}
\caption{Comparison with the state-of-the-art on Something-Something V1. The gray rows represent that they use two stream method and can not directly compare with our method.}
\label{tab:state_of_the_sthv1}
\end{table*}

\begin{table}[t]
\centering

\begin{tabular}{c|c|c}
\toprule[2pt]
Method &Val  & Test \\
\hline
TSN$_{16f\times10}$~\cite{tsn} & 30.0\% &-\\
\hline
TRN-RGB$_{8f}$~\cite{trn} & 48.8\% &50.9\%\\
\color{mygray}TRN-2Stream$_{8f}$~\cite{trn} & \color{mygray}55.5\% &\color{mygray}83.1\%\\
\hline
TSM-RGB$_{8f \times 10}$~(Lin et al. 2018)& 59.1\% & - \\
TSM-RGB$_{16f \times 10}$~(Lin et al. 2018) & 59.4\% & 60.4\% \\
\color{mygray}TSM-2Stream$_{16f\times 10}$~(Lin et al. 2018) & \color{mygray}64.0\% & \color{mygray}64.3\% \\
\hline
TEINet-RGB$_{8f}$ & 61.3\% & -\\
TEINet-RGB$_{8f\times 10}$ & 64.0\% & 62.7\%\\
TEINet-RGB$_{16f}$ & 62.1\% & -\\
TEINet RGB$_{16f\times 10}$ & 64.7\% & 63.0\%\\
TEINet RGB$_{16f+8f}$ & \textbf{66.5\%} & \textbf{64.6\%}\\
\bottomrule[2pt]

\end{tabular}
\caption{Comparison with the state-of-the-art on Something-Something V2. The subscript $8f\times 10$ denotes we sample 10 clips and each clip contain 8 frames.}
\label{tab:state_of_the_sthv2}
\end{table}

\begin{table*}[th]
\centering
\scalebox{0.9}{
\begin{tabular}{ccccccc}
\toprule[2pt]
Method & Backbone & Pre-train & GFLOPs$\times$views & Top-1 & Top-5\\
\hline
I3D$_{64f}$~(Carreira et al. 2017) & Inception V1 & ImgNet & 108$\times$N/A & 72.1\% & 90.3\%\\
I3D$_{64f}$+TSN~\cite{X00LTG19} & Inception V1 & ImgNet & 108$\times$N/A & 73.5\% & 91.6\%\\
ARTNet$_{16f}$+TSN~\cite{artnet} & ResNet-18 & From Scratch & 23.5$\times$250 & 70.7\% & 89.3\%\\
NL+I3D$_{32f}$~\cite{nonlocal} & ResNet-50 & ImgNet & 70.5$\times$30 & 74.9\% & 91.6\% \\
NL+I3D$_{128f}$~\cite{nonlocal} & ResNet-101 & ImgNet & 359$\times$30 & 77.7\% & 93.3\% \\
Slowfast~\cite{slowfast} & ResNet-50 & From Scratch & 36.1$\times$30 & 75.6\% & 92.1\% \\
Slowfast~\cite{slowfast} & ResNet-101 & From Scratch & 106$\times$30 & 77.9\% & 93.2\% \\
NL+Slowfast~\cite{slowfast} & ResNet-101 & From Scratch & 234$\times$30 & \textbf{79.8\%} & \textbf{93.9\%} \\
LGD-3D$_{128f}$~\cite{lgdnet} & ResNet-101 & ImgNet & N/A$\times$N/A & 79.4\% & 94.4\% \\
\hline
TSN~\cite{tsn} & Inception V3 & ImgNet  & 3.2$\times$250 & 72.5\% & 90.2\% \\
ECO$_{En}$~\cite{eco} & BNIncep+Res3D-18 & From Scratch & N/A$\times$N/A & 70.7\% & 89.4\%\\
R(2+1)D$_{32f}$~\cite{r(2+1)d} & ResNet-34 & Sports-1M & 152$\times$10 & 74.3\% & 91.4\%\\

S3D-G$_{64f}$~\cite{s3d} & Inception V1 & ImgNet & 71.4$\times$30 & 74.7\% & 93.4\% \\
StNet$_{25f}$~\cite{stnet} & ResNet-50 & ImgNet & 189.3$\times$1 & 69.9\% & -\\
TSM$_{16f}$~\cite{tsm} & ResNet-50 & ImgNet & 65$\times$30 & 74.7\% & 91.4\%\\
\hline
TEINet$_{8f}$ & ResNet-50& ImgNet & 33$\times$30 & 74.9\% & 91.8\% \\
TEINet$_{16f}$ & ResNet-50& ImgNet & 66$\times$30 & 76.2\% & 92.5\% \\
\bottomrule[2pt]

\end{tabular}
}
\caption{Comparison with the state-of-the-art models on Kinetics-400. Similar to \cite{slowfast}, we report the inference cost by computing the GFLOPs (of a single view) $\times$ the number of views (temporal clips with spatial crops). The subscript $8f$ denotes each clip contains 8-frame and N/A denotes the numbers are not available for us.}
\label{tab:state_of_the_kinetics}
\end{table*}

\begin{table}[th]
\centering
\scalebox{0.9}{
\begin{tabular}{cccc}
\toprule[2pt]
Method  & UCF & HMDB\\
\hline
TSN-RGB~\cite{tsn}  & 93.2\% &-\\
I3D-RGB~(Carreira et al. 2017)  & 95.6\% & 74.8\%\\
P3D-RGB~\cite{p3d} & 88.6\% & - \\
S3D-G-RGB~\cite{s3d} & 96.8\% & 75.9\% \\
R(2+1)D-RGB~\cite{r(2+1)d}  & 96.8\% & 74.5\%\\
TSM-RGB~(Lin et al. 2018) & 96.0\% & 73.2\%\\

ECO$_{En}$~(Zolfaghari et al. 2018)  & 94.8\% & 72.4\%\\

ARTNet-RGB~\cite{artnet}  & 94.3\% & 70.9\%\\
StNet-RGB~\cite{stnet}  & 93.5\% & -\\
\hline
TEINet-RGB & 96.7\% & 72.1\% \\
\bottomrule[2pt]

\end{tabular}
}
\caption{Comparison with the state-of-the-art models on UCF101 and HMDB51. The results are followed common practice that reports accuracy by averaging over all 3 splits. For fair comparison, we only list the models using RGB as inputs.}
\label{tab:state_of_the_ucf_hmdb}

\end{table}

\subsection{Ablation Studies}
\label{ablation_study}
This section provides ablation studies on TEI module design and integration with ResNet50 on the Something-Something V1 dataset. In this section, we report the experimental results using the testing scheme of center crop and one clip, the results are summarized in  Table~\ref{tab:studies_on_TEINet}.

\noindent \textbf{Study on MEM and TIM.}
We first conduct a separate study on the effect of each individual module (MEM or TIM) on action recognition. We find that the TIM is able to yield a better recognition accuracy than MEM (46.1\% vs. 33.5\%), indicating that temporal contextual information is more important for action recognition in the Something-Something dataset. Then, we compare our TIM with other efficient temporal modeling baselines such as TSN and TSM, which demonstrates that our TIM is more effective than these baseline methods. Finally, we compare the performance of MEM with SE on action recognition, and we see that MEM+TIM is better than SE+TIM by 1.3\%, which confirms our motivation that motion-aware attention is better at capturing discriminative temporal features for action recognition.

\noindent \textbf{Which stage to insert TEI blocks.}
As shown in Table~\ref{tab:insert_block}, we find that a clear performance improvement will be obtained when inserting TEI block in the later stages. It is worth noting that res$_4$ has 3 more blocks than res$_5$, but integration at both locations achieves a similar result. The temporal modeling based on higher level features may be more beneficial to recognition, which agrees with the findings from~\cite{s3d}.

\noindent \textbf{The number of TEI block inserted into network.}
Efficiency is an important issue and sometimes we may focus on improving recognition accuracy with a limited extra computation consumption.
We here expect to figure out how many TEI blocks can obtain a trade-off performance. Specifically, We attempt to gradually add TEI blocks from res$_5$ to res$_2$ in ResNet50. As shown in Table~\ref{tab:number_of_block} where we use the same inference settings as Table~\ref{tab:impact_of_two_modules}, 
we can boost the performance by inserting more TEI blocks. 
We also see that res$_{2-5}$ only outperforms res$_{4-5}$ by 0.7\%, but with extra 7 TEI blocks. Therefore, in practice, we recommend to use TEI block simply in stages of res$_{4-5}$ as it is more efficient. But our default choice of the remaining experiments are ready to use the ResNet50 equipped with TEI blocks in all stages.

\noindent \textbf{Analysis on runtime.}
The runtime of model has also drawn considerable attention from researchers in recent years. Several experiments are conducted on Something-Something V1 to manifest the latency and throughput for our models. For the fair comparisons with other models, we follow the inference settings in \cite{tsm} by using a single NVIDIA Tesla P100 GPU to measure the latency and throughput. We use a batch size of 1 to measure the latency and a batch size of 16 to measure the throughput. Data loading time is not considered in this experiment. As shown in Table~\ref{tab:runtime}, Our models achieve the acceptable latency and Throughput comparing with other models.

\subsection{Comparison with the State of the Art}
\label{comp_sota}

\noindent \textbf{Results on Something-Something V1.} 
We compare our TEINet with the current state-of-the-art models in Table~\ref{tab:state_of_the_sthv1}.
It is worth noting that our proposed models are only pre-trained on ImageNet. For fair and detailed comparison, the results of TEINet apply center crop when sampling 1 clip, and 3 crops when sampling 10 clips. We notice that our TEINets dramatically outperform TSN~\cite{tsn}, which demonstrates the effectiveness of TEI Block.
When using 16 frames as input our proposed TEINet outperforms TSM~\cite{tsm} by 5.1\% on validation set and even achieves superior performance to TSM$_{En}$ which ensembles the results of 8 frames and 16 frames. As manifested in Table~\ref{tab:state_of_the_sthv1}, Our TEINet$_{En}$ which has the same setting as TSM$_{En}$ can surpass all existing RGB or RGB+Flow based models on Something-Something V1. When it comes to computational costs, we also list FLOPs for most models, We find that our model achieves the superior performance with reasonable FLOPs during testing. More analyses on runtime comparing with other models have been mentioned in Section~\ref{ablation_study}.

\noindent \textbf{Results on Something-Something V2.} 
As shown in Table~\ref{tab:state_of_the_sthv2}, We report the results on Something-Something V2 which is a new release of V1. The training setting and inference protocol of Table~\ref{tab:state_of_the_sthv2} are consistent with Table~\ref{tab:state_of_the_sthv1}.
Our proposed TEINets obtain the similar performance gain on Something-Something V2 by only using RGB as input. The TEINet$_{8f}$ even achieves 61.3\% and outperforms TSM$_{16f\times10}$ as inputs by 1.9\%. Furthermore, our proposed TEINet$_{16f+8f}$ which ensembles the models using 16 frames and 8 frames as inputs outperforms TSM-2Stream by 2.5\%, and achieves superior performance to the previous state-of-the-art models, which demonstrates that our TEINet is able to capture temporal features on this motion sensitive dataset.

\noindent \textbf{Results on Kinetics-400.}
The Kinetics is currently the most popular dataset in action recognition, due to its large numbers of videos and various categories. 
The results are summarized in Table~\ref{tab:state_of_the_kinetics}.
The upper part of Table~\ref{tab:state_of_the_kinetics} lists the current state-of-the-art models based on 3D convolutions, which are with expensive computational costs. The middle part of Table~\ref{tab:state_of_the_kinetics} lists several slightly lightweight models which are mainly composed of 2D convolutions or a few 3D convolutions in network. Notably, our models based on 2D ResNet-50 only utilize ImageNet as pre-training dataset.
We here only list the models only using RGB as inputs to perform comparisons.
As shown in Table~\ref{tab:state_of_the_kinetics}, our TEINet obtain a better performance gain among lightweight models. Meanwhile, our models even achieve competitive performance when comparing with computationally expensive models. For example, the TEINet using 16-frame as inputs outperforms NL I3D using 32-frame as inputs by 1.3\%.

\noindent \textbf{Transferring to UCF101 and HMDB51.}
To verify the generalization of TEINet on smaller datasets, we evaluate the performance for our models on UCF101 and HMDB51. We fine-tune our TEINet on the UCF10 and HMDB51 datasets using model pre-trained on Kinetics-400, and report the performance using 10 clips and 3 crops per video.
We here only list our model using 16-frame as inputs.
As shown in Table~\ref{tab:state_of_the_ucf_hmdb}, our proposed TEINet also achieves competitive performance when comparing with I3D-RGB and R(2+1)D-RGB, which demonstrates the the generalization ability of our method.

\section{Conclusion}
In this work, we have proposed an efficient temporal modeling method, i.e., TEINet, to capture temporal features in video frames for action recognition. The vanilla ResNet can be converted into TEINet by inserting the TEI blocks which are composed of a motion enhanced module (MEM) and a temporal interaction module (TIM). The MEM focuses on enhancing the motion-related features by calculating temporal attention weights, and TIM is to learn the temporal contextual features with a channel-wise temporal convolution.
We conducted a series of empirical studies to demonstrate the effectiveness of TEINet for action recognition in videos.
The experimental results show that our method has achieved the state-of-the-art performance on the Something-Something V1\&V2 dataset and competitive performance on the Kinetics dataset with a high efficiency.

\section{Acknowledgments}
This work is supported by the National Science Foundation of China (No. 61921006), and Collaborative Innovation Center of Novel Software Technology and Industrialization.

\small\bibliographystyle{aaai}\bibliography{ref}

\end{document}